\documentclass[lettersize,journal]{IEEEtran}
\usepackage{amsmath,amsfonts}
\usepackage{algorithmic}
\usepackage{algorithm}
\usepackage{array}
\usepackage[caption=false,font=normalsize,labelfont=sf,textfont=sf]{subfig}
\usepackage{textcomp}
\usepackage{stfloats}
\usepackage{url}
\usepackage{verbatim}
\usepackage{graphicx}
\usepackage{cite}
\usepackage{booktabs}
\usepackage{amsmath}
\usepackage[hidelinks, colorlinks]{hyperref}
\usepackage{xcolor}

\begin{document}

\title{GeoRSMLLM: A Multimodal Large Language Model for Vision-Language Tasks in Geoscience and Remote Sensing}

\author{Zilun Zhang $\dagger$, 
        Haozhan Shen $\dagger$,
        Tiancheng Zhao,
        Bin Chen,
        Zian Guan, \\
        Yuhao Wang, 
        Xu Jia,
        Yuxiang Cai,
        Yongheng Shang,
        Jianwei Yin
\thanks{$\dagger$: Equal Contribution}

\thanks{
Corresponding author: Tiancheng Zhao. Zilun Zhang, Haozhan Shen, Bin Chen, Zian Guan, Yuhao Wang, Xu Jia, Yuxiang Cai, Yongheng Shang, and Jianwei Yin are with the College of Computer Science and Technology, Zhejiang University, Hangzhou, China; Tiancheng Zhao is with the Binjiang Research Institute of Zhejiang University (e-mail: zilun.zhang@zju.edu.cn; tianchez@zju-bj.com).
}
}

\markboth{Journal of \LaTeX\ Class Files,~Vol.~14, No.~8, August~2021}%
{Shell \MakeLowercase{\textit{et al.}}: A Sample Article Using IEEEtran.cls for IEEE Journals}


\maketitle

\begin{abstract} The application of Vision-Language Models (VLMs) in remote sensing (RS) has demonstrated significant potential in traditional tasks such as scene classification, object detection, and image captioning. However, current models, which excel in Referring Expression Comprehension (REC), struggle with tasks involving complex instructions (e.g., exists multiple conditions) or pixel-level operations like segmentation and change detection. In this white paper, we provide a comprehensive hierarchical summary of vision-language tasks in RS, categorized by the varying levels of cognitive capability required. We introduce the Remote Sensing Vision-Language Task Set (RSVLTS), which includes Open-Vocabulary Tasks (OVT), Referring Expression Tasks (RET), and Described Object Tasks (DOT) with increased difficulty, and Visual Question Answering (VQA) aloneside. Moreover, we propose a novel unified data representation using a set-of-points approach for RSVLTS, along with a condition parser and a self-augmentation strategy based on cyclic referring. These features are integrated into the GeoRSMLLM model, and this enhanced model is designed to handle a broad range of tasks of RSVLTS, paving the way for a more generalized solution for vision-language tasks in geoscience and remote sensing.
\end{abstract}


\section{Introduction}
The advent of deep learning has revolutionized the field of remote sensing (RS) by enabling advanced analysis based on enormous Earth observation data. Many traditional tasks such as scene classification, object detection, semantic segmentation, change detection, object counting, etc. only involve imagery data for vision-only downstream tasks. 
Since 2021, Vision-Language Models (VLMs) that leveraged the rich correlations between visual content and linguistic descriptions, offering an understanding of vision and textual data beyond traditional computer vision techniques \cite{zhang2024visionlanguagemodelsvisiontasks}. 


In general, there are two types of VLMs. \textbf{Contrastive VLMs} \cite{clip} \cite{declip} train visual and text encoders with contrastive loss to set up the shallow association between visual and linguistic data. They are good at Zero-Shot Classification (ZSC) and Image-Text Retrieval. \textbf{Generative VLMs} \cite{albef} \cite{liu2023grounding} can handle more complex tasks such as Visual Question Answering (VQA) and Visual Grounding (VG). With the rise of ChatGPT and other Large Language Models (LLMs) in 2022, Multimodal Large Language Models (MLLMs) like GPT-4V and LLaVA show the capabilities to process and reason over multiple modalities, such as text, images \cite{liu2023llava} \cite{minigptv2}, audio, and even depth \cite{girdhar2023imagebindembeddingspacebind}. \textbf{When MLLMs are designed for only two modalities - vision and language, they can be seen as a special case of VLMs} (When we mentioned the "MLLMs" in this paper, we specifically refer to VLMs that use LLMs as base model). Different from Contrastive VLMs that aim to build the shallow association between visual data and linguistic data, these models are designed to tackle a wide array of tasks that require a complex comprehensive understanding of both linguistic and visual information with the help of LLMs. Ultimately, they are designed to respond the complex instructions like visual assistants \cite{yin2024surveymultimodallargelanguage}.

The potential of VLMs for RS has been widely studied since 2023. Li et al. \cite{li2024visionlanguagemodelsremotesensing} provide a comprehensive review of the current progress and future trends of VLMs in RS, highlighting the capabilities in tasks such as image captioning, VQA, and object detection. Emphasizing the potential of these models for deeper semantic understanding in RS applications. Zhou et al. \cite{zhou2024visionlanguagegeofoundationmodelsurvey} examined the latest progress in VLMs for RS, and taxonomized the data pipeline, model architecture, and capability for downstream tasks from recent works. Many of them have demonstrated remarkable progress by providing intelligent solutions that closely mimic human decision-making based on visual information and textual instruction. However, current models have limitations such as solving complex reasoning tasks with multiple conditions, as mentioned in section \ref{sec:diff}.

The contributions of our work are threefold. We provide \textbf{a hierarchical summary of tasks for VLMs in RS} based on the different levels of cognition capability required by tasks. The limitations of current VLMs in RS are discussed. Then we introduce \textbf{a data representation scheme that can unify all tasks} and a condition parser module, alone with model design. Once the data representation for different tasks is unified, only model structures need to be improved in the future. The full paper follows this with deeper explanations, more experiment results and comparison, as discussed in section \ref{sec:outline}.


\section{Remote Sensing Vision-Language Task Set and Described Object Task}
\label{sec:taxonomy}

\begin{figure}[ht]
    \centering
    \includegraphics[width=0.5\textwidth]{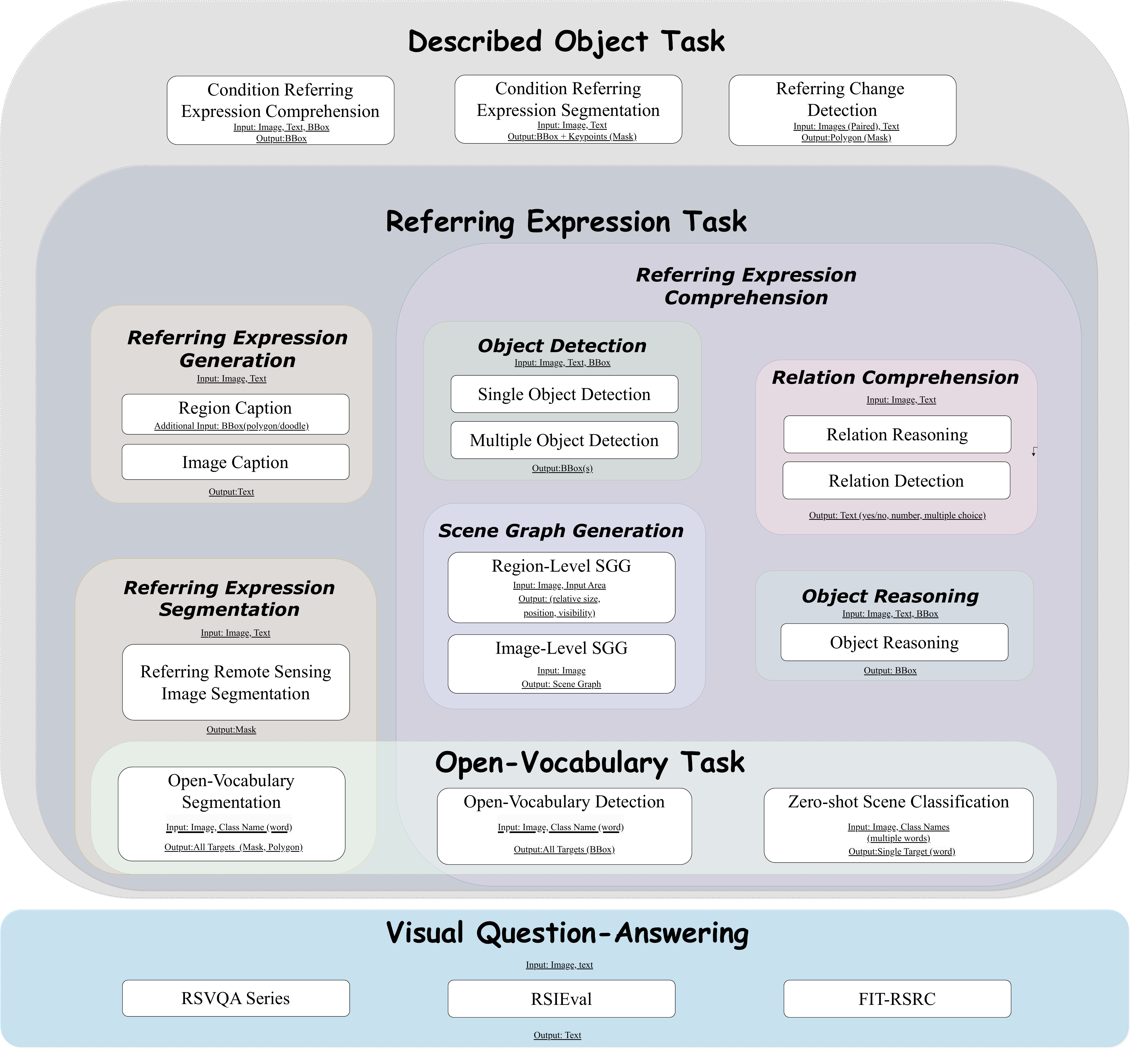}
    \caption{Hierarchical summary of Vision-Language Task in RS and Remote Sensing Vision-Language Task Set (RSVLTS). This task set includes \textbf{Open-Vocabulary Task} (OVT), \textbf{Referring Expression Task} (RET), \textbf{Described Object Task} (DOT, contains the former two) and \textbf{Visual Question Answering} (VQA).  
    Typical inputs and outputs for each task are listed. This figure omits details for each task, which will be discussed thoroughly in section \ref{sec:taxonomy}. 
    }
    \label{fig:DOTS}
\end{figure}

 We begin by categorizing the VLM tasks in RS. Based on this hierarchical summary, we introduce the Remote Sensing Vision-Language Task Set (RSVLTS). Inspired by Xie et al.'s Described Object Detection (DOD) concept \cite{xie2023DOD}, RSVLTS consists of three progressively challenging sub-tasks: 	\textbf{Open-Vocabulary Task} (OVT), 	\textbf{Referring Expression Task} (RET), and 	\textbf{Described Object Task} (DOT), each requiring increasing levels of cognitive ability. Models capable of completing more complex tasks are inherently able to handle simpler ones. It is worth mentioning that some VQA tasks do not refer to any specific object, so we categorize them alongside DOT within the RSVLTS. Figure \ref{fig:DOTS} illustrates the hierarchical summary of Vision-Language tasks in RS, along with details of the RSVLTS. Text-to-image task is also a vision-langugae task, but we decide to not expand it here.

\subsection{Open-Vocabulary Task}

The Open-Vocabulary Task refers to a set of tasks where models are required to recognize and understand a vast array of objects, scenes, or concepts that extend beyond a predefined set of categories \cite{wu2024openvocabularylearningsurvey}. This is in contrast to traditional closed-set tasks, where models are only expected to recognize a limited, predefined set of classes (e.g., ImageNet and COCO competition). The input can be an image and text, and the output can include text, multiple bounding boxes, or segmentation masks. Open-Vocabulary tasks closely resemble real-world scenarios where models must generalize and adapt to new, unseen classes that were not present in the training data \cite{zhu2024surveyopenvocabularydetectionsegmentation}. Such models can adapt to new categories without the need for retraining, making them more flexible and suitable for dynamic environments. Additionally, for visually similar concepts (e.g., lake and river), closed-vocabulary models may struggle due to different class labels assigned during pre-training, whereas open-vocabulary models handle them seamlessly.


There are three major types of Open-Vocabulary Tasks (OVT) in RS. \textbf{Zero-shot Classification} leverages the robust image-text correlation capabilities of VLMs. The core objective is to classify an image into the most appropriate category by ranking the similarity between text and visual features (e.g., GeoRSCLIP \cite{rs5m}). This approach allows VLMs to transform and solve traditional scene classification tasks in RS. \textbf{Open-Vocabulary Detection and Segmentation} enable the detection or segmentation of new, unseen concepts. The task inputs usually consist of a class name and an image, while the outputs are bounding boxes (e.g., LAE-DINO \cite{pan2024locateearthadvancingopenvocabulary}) or segmentation masks (e.g., SegEarth-ov \cite{li2024segearthovtraningfreeopenvocabularysegmentation} and SeLo \cite{selo}) for all targets belonging to that class in the remote sensing image. Many works utilize CLIP's inherent localization capabilities or the power of foundation models  \cite{ravi2024sam2}.

\subsection{Referring Expression Task}

Referring Expression Tasks (RET) in the field of RS can be divided into three major categories: 	\textbf{Referring Expression Comprehension} (REC), 	\textbf{Referring Expression Segmentation} (RES), and 	\textbf{Referring Expression Generation} (REG). REC and RES are closely linked tasks that share the common goal of interpreting a linguistic expression to localize the corresponding entity or region within a visual scene \cite{qiao2020referringexpressioncomprehensionsurvey}. REG takes an image and a text instruction as input and outputs the image caption (\textbf{Image Caption} task) or region caption if additional region information is provided (\textbf{Region Caption} task). REC can be divided into several sub-tasks. \textbf{Object Detection} requires the model to detect one or more bounding boxes as specified by the input text instruction, while \textbf{Object Reasoning} requires the model to output the categories and locations of all targets that meet the conditions for a given specific relation and random categories and locations as the query. \textbf{Visual Grounding}, mentioned by GeoChat \cite{geochat}, EarthGPT \cite{earthgpt}, and H2RSVLM \cite{H2RSVLM}, has slightly different definitions among these works. We formalize visual grounding as \textbf{Object Detection} task (under REC circumstances) when one or multiple objects need to be localized (outputting coordinates of bounding boxes) based on the given text input, and as a \textbf{Region Caption} (REG) task when the caption needs to be provided with a specified location. \textbf{Visual Question Answering} (VQA) is also a sub-task of REC, where, given an image and a text question, an answer is required. This answer could be "Yes/No" (e.g., Presence, Comparison, Rural/Urban, Existence), a Number (e.g., counting), or Multiple Choice (e.g., Relation Detection, Relation Reasoning, Relative Position, Color, Road Direction). While the question may require complex logical reasoning, the answer itself is often straightforward and could potentially be exploited by shortcuts within the model since the reasoning steps are not required. VRSBench \cite{vrsbench} addresses this issue by using GPT-4 to evaluate the quality of the answer along with the logical steps taken to derive it.

MLLM models with autoregressive decoder-only structures have become a common choice for REC and REG. GeoChat \cite{geochat}, EarthGPT \cite{earthgpt}, SkysenseGPT \cite{skysensegpt}, and H2RSVLM \cite{H2RSVLM} effectively solve REC and REG tasks by providing appropriate responses containing coordinate information  or caption after instruction tuning with geoscience and RS-related data. However, there is no unified model that can do all RET.


The key difference between RET and OVT is the required degree of the model's capability to understand the given text instructions. For OVT, only weak associations between visual and textual information are needed, and the text input is generally a single phrase. As the majority choice for OVT, CLIP-based approaches has the maximum text context length of 77 tokens, with an much shorter effective context length \cite{zhang2024longclipunlockinglongtextcapability}. CLIP-based models only need to produce high similarity scores when visual and text features are close in the embedding space and vice versa, rather than deeply understanding the text meaning. As a result, most OVT models struggle with complex text instructions for image comprehension. Models capable of handling RET can also handle OVT, although there may be a trade-off between inference speed and capability.

\subsection{Described Object Task}
\label{sec:diff}

There are several limitations for models designed for RET. First, they struggle with complex instructions involving multiple conditions. Second, while solutions for REG and REC can be unified using MLLMs, these models are unable to handle RES, which often requires specialized models like RMSIN \cite{liu2024rotatedmultiscaleinteractionnetwork}. Third, current MLLMs are less effective at traditional RS tasks like change detection, which involves analyzing multiple input images and generating pixel-level binary change masks. 

To emphasize these limitations, we introduce a more advanced and challenging task than RET, which we call the Described Object Task (DOT). This includes Conditional Referring Expression Comprehension (CREC), Conditional Referring Expression Segmentation (CRES), Referring Change Detection (RCD, a text instruction guides the model to focus on changes in a specific region), and more related tasks. 

\begin{figure}[ht]
    \centering
    \includegraphics[width=0.5\textwidth]{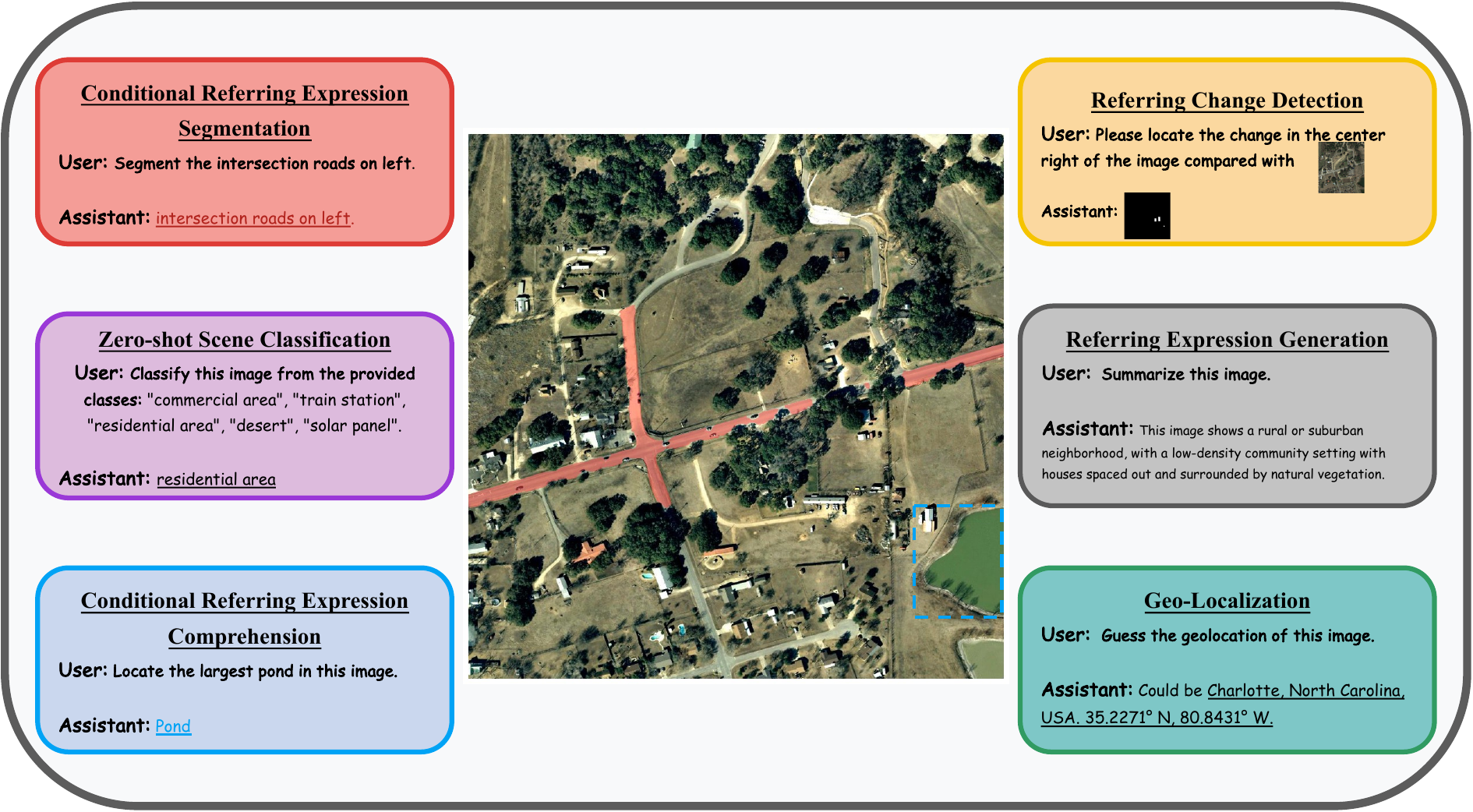}
    \caption{Example of sub-tasks from RSVLTS. Conditional Referring Expression Comprehension, Conditional Referring Expression Segmentation, Referring Change Detection, Referring Expression Generation, Zero-shot Scene Classification, and Geo-localization are shown.
    }
    \label{fig:georsmllm_app}
\end{figure}

A model capable of solving DOT tasks can seamlessly handle both OVT and RET. To this end, we have developed a new data representation scheme and a unified model capable of addressing all DOT and various other related RS challenges. Figure \ref{fig:georsmllm_app} illustrates task examples for the proposed RSVLTS, emphasizing on DOT. Benchmarks to evaluate DOT will be constructed based on existing  REC, RES, and Change Detection dataset.

\section{Unified Model for RSVLTS}
There are 3 types of approaches for MLLM to do the downstream tasks based on the encode and decode approaches. 

LLaVA-like models \cite{liu2023improvedllava}, \cite{geochat}, \cite{skysensegpt} enhance inputs by adding additional data like location coordinates, which are encoded alongside textual instructions to the word embedding of the LLM. Task types, such as "[grounding]" or "[detection]," are indicated at the beginning of the instruction. The model generates output tokens containing the required information, such as location coordinates, in natural text. The specific output format is learned implicitly through supervised fine-tuning \textbf{without introducing extra tokens or decoders.}

The pixel-to-sequence paradigm, demonstrated by models like VisionLLM \cite{wang2023visionllmlargelanguagemodel} and LISA \cite{lai2024lisareasoningsegmentationlarge}, encodes both text and other inputs together into word embeddings, which are then processed by the LLM. In this approach, models generate \textbf{special tokens} for downstream tasks (e.g., detection, segmentation) that are decoded using \textbf{task-specific decoders.} 

In contrast, the pixel-to-embedding paradigm, proposed by NExT-Chat \cite{nextchat}, \textbf{encodes location data into an embedding} using specific encoder along with word embeddings, feeding them together into the LLM. The model then outputs a \textbf{shared trigger token}, which is passed to a box decoder and mask head for \textbf{task-specific decoding}.

Introducing new tokens to learn often increases the complexity of training and limits the model’s ability to adapt to new downstream tasks without retraining. Successful works like Geochat, SkysenseGPT, and VRSBench have shown that organizing location information directly into text instructions is an effective approach. We aim to extend these approaches to tasks such as segmentation and change detection.

In this section, we use $F$ to represent the MLLM, $x_{\text{text}} \in \mathbb{R}^{n_1 \times d}$ for input text embedding, and $x_{\text{image}} \in \mathbb{R}^{n_2 \times d}$ for input image embedding, where $n_1$ is the context length for input text, $n_2$ is the number of tokens to represent an image, and $d$ is the dimension of embedding.

\subsection{Unified Data Representation}

We use the \textbf{set-of-point} representation to unify the inputs across various downstream tasks. 

\subsubsection{Object Detection}  




Geochat, EarthGPT, and SkysenseGPT use $[cx, cy, w, h, \theta]$ to represent the output rotational bounding box, where $cx$ and $cy$ are the center point coordinates, $w$ and $h$ are width and height, and $\theta$ is the rotation angle. VRSBench suggests that $[x1, y1, x2, y2, \theta]$, which uses the coordinates of the top-left and bottom-right points, can offer better performance. After experimental validation, we selected \textbf{the coordinates of four corner points}, $[(x1, y1), (x2, y2), (x3, y3), (x4, y4)]$, as the representation for rotational bounding boxes. The object detection task can be formalized as follows:

\begin{align}
    \{rbb_i\}_{i=1}^k = F(x_{\text{image}}, x_{\text{text}})
\end{align}

\noindent where $rbb_i = [(x1, y1), (x2, y2), (x3, y3), (x4, y4)]_i$ represents the corner coordinates of the $i^{th}$ detected target, and $k$ denotes the number of targets that satisfy the given instruction. In this task, we simply convert the data from the training and validation set of SkysenseGPT and VRSBench that related to object detection tasks.

\subsubsection{Segmentation} 
The pixel-wise segmentation masks $\{m_i\}_{i=1}^{k}$ are produced with the aid of SAM2 \cite{ravi2024sam2} (denote as $S$) using \textbf{box and keypoints prompts} in a zero-shot manner. The segmentation task RS can be formalized as follows:
\begin{align}
    \{hbb_i, \{p_{i}^{j}\}_{j=1}^n\}_{i=1}^k &= F(x_{\text{image}}, x_{\text{text}}) \\
    \{m_i\}_{i=1}^k &= S(x_{\text{image}}, hbb_i, \{p_{i}^{j}\}_{j=1}^n\})_{i=1}^k
\end{align}

\noindent where $hbb_i = [(x1, y1), (x2, y2)]_i$ represents the \textbf{coordinates of the top-left and bottom-right points of the bounding box}, and $p_i = [(x1, y1), \cdots, (xn, yn)]_i$ represents the coordinates of keypoints within such bounding box of the $i^{th}$ target, and $k$ denotes the number of targets.
We converted RemoteContour34K \cite{miao2024promptingdirectsamsemanticcontour} dataset to bounding boxes and keypoints to instruction tuning data, which includes data from LoveDA \cite{loveda}, iSAID \cite{isaid}, DeepGlobe \cite{8575485}, and RefSegRS \cite{yuan2024rrsisreferringremotesensing} datasets.

\subsubsection{Referring Change Detection}
The mask for change is represented using \textbf{polygons}.
The referring change detection task can be formalized as follows:
\begin{align}
    \{poly_i\}_{i=1}^k = F(x_{\text{imageA}}, x_{\text{imageB}}, x_{\text{text}})
\end{align}

\noindent where $poly_i = [(x1, y1), \cdots, (xn, yn), (x1, y1)]_i$ represents the coordinates of a set of points forming a closed polygon (with the first and last points overlapping), and $k$ denotes the number of polygons. These polygons can later be converted into binary masks. The motivation behind using polygon rather than bounding box and keypoints is that change detection targets are usually much smaller than segmentation targets. The instruction tuning data for this task is derived from the LEVIR-MCI dataset \cite{Liu_2024}, which includes 10K bi-temporal images, each annotated with corresponding masks and five captions describing the changes in each image.

\subsubsection{Other Tasks} There are many more geoscience and remote sensing tasks that can be integrated into RSVLTS. For example, Object counting \cite{zhang2024goodcaptioningbadcounting} and geolocalization \cite{cepeda2023geoclipclipinspiredalignmentlocations}. Taking geolocalization as an example, it aims to pinpoint the precise location of images taken anywhere on Earth. Such geolocalization task can be formalized as follows:

\begin{align}
    [\text{city name}, (\text{latitude}, \text{longitude})] = F(x_{\text{image}}, x_{\text{text}})
\end{align}


Following pioneer works \cite{geochat} \cite{skysensegpt}, the task notation such as "[seg]", "[change]", "[geoloc]", etc. are included in the beginning of each instruction.
 
\subsection{Condition Parser}



To address the Conditional Referring Expression Comprehension and Segmentation tasks, an instruction parser is required. Given an instruction \( c \) with multiple conditions and a parser \( \mathbf{P} \), the instruction can be broken down into a set of sub-instructions \( c_i \) each containing one condition. Then the MLLM solves iteratively under the chain-of-thought scheme.

\begin{align}
    \{c_i\}_{i=1}^k = \mathbf{P}(c)
\end{align}

For example, the instruction "detect all planes on the east bank of the river" would be broken into two sub-instructions. First, the MLLM executes "detect all planes" and obtains a response. Then, it proceeds to "select the one on the east bank of the river," using the result of the previous sub-instruction. The parser $P$ is implemented by an independent LLM.

\subsection{Self-Augmentation by Cyclic Referring}



There are fewer than 50K instructions for the RES and RCD tasks. To help the model become more familiar with the set-of-point representation, we convert the REC, RES, and RCD tasks into a region captioning task. The motivation behind this is that the outputs of REC, RES, and RCD tasks can serve as inputs for the region captioning task with only slight modifications to the prompts, and vice versa. For instance, a simple pseudo example for REC and REG:
\begin{quote}
\small
\texttt{Q: "What's the location of the largest pond in this image?"} \\
\texttt{A: \{(100, 100), (100, 200), (200, 200), (200, 100)\}} \\[5pt]
\texttt{Q: "Could you describe the object at \{(100, 100), (100, 200), (200, 200), (200, 100)\}?"} \\
\texttt{A: A large pond}
\end{quote}

\subsection{Model Detail}
The main model is based on the LLaVA structure \cite{liu2023llava}, with several key modifications. First, the LLM backbone is replaced with Qwen2-7B, as the set-of-points representation requires support for much longer context. Additionally, the vision tower from SigLIP \cite{zhai2023sigmoidlosslanguageimage} is used. To handle high-resolution images, we apply the dynamic high-resolution scheme technique \cite{liu2024llavanext}. The independent condition parser block is added.

\newpage

\newpage
\bibliographystyle{IEEEtran}
\bibliography{ref}

\end{document}